\documentclass[11pt,a4paper]{article}
\usepackage{acl2017}
\usepackage{times}
\usepackage{latexsym}

\usepackage{amsmath}
\usepackage{amsfonts,amssymb}
\usepackage{graphicx}
\usepackage{multirow}
\usepackage{wrapfig}
\usepackage{epstopdf}

\aclfinalcopy 

\title{Online Deception Detection Refueled by Real World Data Collection}

\author{Wenlin Yao, Zeyu Dai, Ruihong Huang, James Caverlee \\
        Department of Computer Science and Engineering\\
		Texas A\&M University\\
         {\tt \{wenlinyao, jzdaizeyu, huangrh, caverlee\}@tamu.edu}}

\date{}

\begin{document}

\maketitle

\begin{abstract}
The lack of large realistic datasets presents a bottleneck in online deception detection studies.
In this paper, we apply a data collection method based on social network analysis to quickly identify high-quality deceptive and truthful online reviews\footnote{Dataset will also be released} from Amazon.
The dataset contains more than 10,000 deceptive reviews and is diverse in product domains and reviewers. 
 Using this dataset, we explore effective general features for online deception detection that perform well across domains.
We demonstrate that with generalized features -- advertising speak and writing complexity scores -- deception detection performance can be further improved by adding additional deceptive reviews from assorted domains in training. Finally, reviewer level evaluation gives an interesting insight into different deceptive reviewers' writing styles.

\end{abstract}

\section{Introduction}

Online reviews are increasingly being used by consumers in making purchase decisions.
A recent survey by the Nielsen Company shows that 57\% of electronic shoppers and 45\% of car shoppers were influenced by online reviews. 
However, due to the widespread growth of crowdsourcing platforms like Mechanical Turk, large-scale organized campaigns 
can be quickly launched and create massive malicious reviews in order to promote products or to defame competitors. Deceptive product reviews can easily bias and mislead consumers' perception of product quality.

Deceptive opinion spam detection is challenging because deception makers can target various objects and domains (e.g., commercial products or services) and the language is dynamic 
to adapt to distinct objects. Basically, the goal of deception detection is to recognize 
varied and generalized
linguistic cues that can indicate deceit across domains, which is dramatically different from semantic analysis (e.g. sentiment analysis).
So far, understanding of deceptive language in general is still scarce. 
One major obstacle is that it is difficult to obtain ground truth deceptive and authentic reviews. 
\citet{ott2011finding} revealed that it is impossible for humans to accurately identify and label deception. 
Therefore, a common 
method to have ground truth labels for deception detection~\cite{li2014towards} is to ask customers or domain experts to write down deceptive reviews following a carefully designed and strictly controlled procedure. Such a data collection approach is slow, costly and hard to scale. Consequently, most of the prior research has used small datasets containing several hundred of reviews that usually take weeks to collect~\cite{ott2011finding}. Moreover, this data collection method is too strict and cannot reflect real world online review manipulation processes. 

Due to the difficulty of deception data collection, most research has been limited to study deception within an individual domain such as hotel or restaurant~\cite{ott2011finding,feng2012syntactic,ott2013negative,li2014towards}.
Consequently, only several conceptual types of features have been studied for deception detection including genre indicative Part-Of-Speech (POS) features and psychologically motivated Linguistic Inquiry and Word Count (LIWC)~\cite{pennebaker2015development} category features. What is surprising is that among all the studied features, Bag-Of-Word (BOW) features remain the most effective type of features for deception detection. Until very recently, \citet{li2014towards} suggested that POS features and LIWC features are better than BOW features in generalizing across domains, solely based on their experiments with several hundred reviews and three domains.
Without large scale realistic data, it is difficult to gain a deeper understanding of general rules for online deception detection.

In this paper, we apply an existing online deception data collection method~\cite{fayazi2015uncovering} that recognizes online campaigns of malicious review posting and further employ social network analysis on reviewer-reviewer graphs, which can quickly accumulate the list of deceptive reviewers as well as deceptive reviews. The deceptive reviews collected in this manner are rich in terms of authors and domains. We constructed a dataset containing around 10,000 deceptive reviews written by 1,540 deceptive reviewers, ranging over more than 30 domains including books, electronics, movies, etc. Using such an author and domain diversified dataset, we are able to conduct extensive cross-domain experiments in order to search for general rules on deception detection. Then we demonstrate that with generalized sets of features, increasing amounts of data from arbitrarily different domains can continuously improve the performance of deception detection, which also shows the value of this 
scalable deception data collection method.
In addition, our preliminary experiments on reviewer level deception analysis shows that detection systems trained with data from reviewers of a type might perform poorly in detecting opinion spam written by reviewers of a contrasting type. 

This paper has two major contributions. First, we connect two communities -- social media analysis and language analysis -- that have worked on online deception detection but with a different focus. We apply an existing social network analysis based algorithm for collecting a large and rich dataset from the wild online world. 
Second, we are the first to conduct linguistically motivated deception detection analysis using such a diverse dataset that is two orders of magnitude larger than the datasets used in  previous studies.
Most prior studies are based on manually collected unreal and small datasets, and the lack of large and real datasets significantly limits the depth of deception detection research.
Our experiments show that general rules for deception detection are likely to be revealed by large-scale cross-domain experiments in modeling writing styles instead of modeling contents as in single-domain experiments.

\section{Related Work}
Previous studies mainly relied on four types of methods to obtain deceptive reviews.
The first method is to ask human annotators to label deceptive reviews~\cite{jindal2010finding, mukherjee2012spotting}. However, studies show that humans are not good at identifying deceptive reviews and the annotation performance is poor~\cite{ott2011finding}. 
Therefore, \citet{ott2011finding,ott2013negative} introduced the second approach that they asked Amazon Mechanical Turkers to compose 400 deceptive and 400 non-deceptive reviews. 
Similarly, \citet{li2014towards} asked both Mechanical Turkers and domain experts to write deceptive reviews.
There are two main drawbacks of this method.
First, it is too strict to scale.
Second, this strict method does not reflect how deceptive reviews are posted in the real online environment.

The third approach used rule-based heuristics to filter reviews which are likely to be deceptive~\cite{jindal2007review, jindal2008opinion, hammad2013approach}. The rules include labeling reviews that contain certain keywords as deceptive, labeling duplicate or near duplicate reviews as deceptive, labeling product irrelevant reviews as deceptive, etc.
Nonetheless, such methods can be easily fooled by careful deceptive review creators. 
Lastly, the fourth method~\cite{mukherjee2013yelp} proposed a deceptive review dataset by 
simply collecting reviews filtered by the Yelp website as ground truth deceptive reviews and collecting reviews not filtered as authentic reviews.
However, the filtering mechanism used by Yelp is still a black box and unknown. Therefore, the data collected by such method is not trustworthy.


So far, due to the lack of a large and realistic dataset, 
language analysis research for online deception detection are restricted.
\citet{jindal2008opinion} first studied the problem of deceptive opinion spam.
They discussed the evolution of opinion mining, which focused on 
summarizing the opinions from text in order to identify duplicate opinions as spam. \citet{ott2011finding} 
presented three different types of basic features for deceptive spam detection including n-grams, POS tags and LIWC. 
\citet{feng2012syntactic} 
found that syntactic stylometry features derived from Context Free Grammar (CFG) parse trees improved deception detection performance over several baselines that use shallow lexico-syntactic features. \citet{li2014towards}
found that LIWC and POS are more robust than n-grams features when applied to cross-domain adaptation. More recently, a multi-task learning method \cite{hai2016deceptive} for deception detection is developed to exploit domain relatedness and to use unlabeled data.


\section{Dataset Construction}
Crowdsourcing platforms have been shown their effectiveness in organizing large amounts of individuals to work on a target task. These platforms also provide an alternative strategy 
to request deceptive reviews in order to promote their own products. Recently, \citet{fayazi2015uncovering} introduced an approach to understand massive manipulation of online reviews. 
For our purpose, we employed their approach to collect real world manipulation of online reviews. 

\begin{figure}[h]
 \centering
 \includegraphics[width = 2.6in]{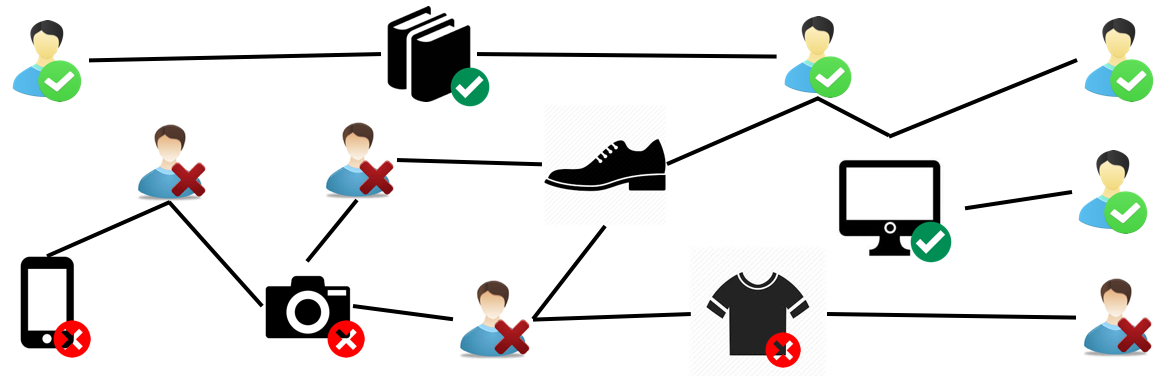}
 \caption{Online Deceptive Review Collection using Social Network Analysis}
\label{social_analysis}
\end{figure}

Briefly, this deceptive review collection approach has two steps. First, we identified deceptive review request tasks from crowdsourcing platforms. Tracking the Amazon URL in each task which links to a specific product, we collected 
reviews associated with the product as well as reviewers' information. We next augmented this base set of reviews, reviewers and products via a three hop breadth-first search, 
and applied social network analysis techniques to a rich product-reviewer-review graph in order to identify additional deceptive review composers and their written deceptive reviews (Figure \ref{social_analysis}). Details are given in section \ref{initial_section} and section \ref{discovering_section}.

\subsection{Initial Products, Reviewers and Reviews}\label{initial_section}
First, we followed the framework presented by \cite{fayazi2015uncovering} and 
crawled deceptive review creation tasks posted on ShortTask.com, RapidWorkers.com, and Microworkers.com. Starting from this as {\it root task set}, we collect initial deceptive  products, reviews and reviewers. Then we crawled three hops to identify a larger candidate set of potential deceptive reviews, reviewers and products.

\subsection{Discovering Additional Deceptive Reviewers}\label{discovering_section}
In order to identify additional deceptive reviewers who have contributed deceptive reviews,  
we applied 
a reviewer-reviewer graph clustering algorithm using a pairwise Markov Random Field (MRF) that defines individual and pair potentials~\cite{fayazi2015uncovering}. Specifically, every node in the random field corresponds to one reviewer. 
Individual level potential (single reviewer) and pairwise potential (between two reviewers) functions are used to capture two intuitions. 
(i) Two reviewers who collaborated (responded to the same task) heavily should be assigned to the same cluster. (ii) Two reviewers who behave similarly in posting reviews probably belong to the same cluster (deceptive or authentic). For instance, one reviewer who actually purchased a product and another reviewer who did not purchase the product should be put to different clusters.

Then, we formalized the goal of identifying additional deception making reviewers using a maximizing the likelihood function, which 
is optimized using an Expectation Maximization (EM) algorithm. EM iterates over two steps to increase the overall likelihood. In specific, E-Step finds the best cluster assignments given current parameters and M-Step updates parameters given the best cluster assignment. 

\subsection{Ground Truth}

The aforementioned social network analysis approach also identifies various deceptive reviewers and deceptive products.
To get ground truth labels of reviews, deceptive or authentic, we applied multiple sieves on reviews, based on their reviewers and targeted products.
Specifically, we deem a review as deceptive if it satisfies the following two conditions:  
(i) Its author is marked as a deceptive reviewer;  
and (ii) The  product this review commented on appears in an initial deceptive review request task. 
Meanwhile, we deem a review as authentic if neither its reviewer nor its commented   product has been labeled as deceptive by the data collection system
\footnote{All deceptive and authentic reviews were collected from 3-hop breadth-first search over the social network graph, as described in Section \ref{initial_section}.}. 

This method of collecting deceptive reviews and authentic reviews is reasonable. First, \citet{fayazi2015uncovering} reported that their social analysis model can achieve high performance - 0.96 AUC (Area Under the ROC Curve) on balanced training and test sets and 0.77 AUC on unbalanced sets\footnote{In their evaluation, the unbalanced training and test sets each contains half of the original data, so real world deceptive to authentic ratio can be kept. Balanced sets undersample authentic reviews so that each set has equal number of deceptive and authentic reviews.}. Second, our double filtering (reviewer-level and product-level) further controls the quality in labeling deceptive and authentic reviews. 

\section{Dataset Overview}
The dataset we collected contains more than 10,000 deceptive reviews, which is significantly larger than the datasets examined in previous studies. In addition, the dataset is from the real world manipulation and is diverse in reviewers and products (Table \ref{dec_dist} shows dataset statistics).
\begin{table}[h]
\small
\begin{center}
\begin{tabular}{ |l|ll| }
  \hline
   & deceptive & truthful \\
   \hline
  reviews &  10114 & 101226 \\
  reviewers & 1540 & 16497 \\
  products & 994 & 72266\\
  \hline
\end{tabular}
\end{center}
\caption{Number of Reviews, Reviewers and Products}
\label{dec_dist}
\end{table}

The deceptive products that were targeted in crowdsourcing websites can be organized into 32 domains based on the product hierarchy on Amazon. Therefore, it is impossible to train a deception detection system for each domain, which motivates our research on deriving general rules for deception detection that can apply across domains. We further see that the domains are highly unbalanced in size, for instance, 4,555 deceptive reviews were identified for ``Kindle Edition'' books while only 38 deceptive reviews were identified for ``Kitchen'' related products, which reflects skewed distribution of deceptive reviews in reality to certain degree. In contrast, the datasets used in prior studies are small and roughly balanced.

\section{General Linguistic Features}

We first discuss 
linguistic features that have been used in previous studies (section \ref{basic_features_section}). 
Then we introduce 
our new types of features in order to model advertising language that is commonly used in deceptive reviews (section \ref{ad_section}) and syntactic complexities (section \ref{syn_comp_section}).

\subsection{Basic Features}\label{basic_features_section}
\label{basicF}
\noindent{\bf N-gram Based Lexical Features}: 
Both unigrams and bigrams \cite{brown1992class} features have been used for deception detection~\cite{ott2011finding}. 
To be consistent with the cross-domain experimental settings in \citet{li2014towards}, we only consider unigram lexical features (BOWs) in our experiment.

\vspace{.05in}
\noindent{\bf LIWC Features}: Features derived using the Linguistic Inquiry and Word Count (LIWC) lexicon~\cite{pennebaker2015development} has been shown effective in both within-domain~\cite{ott2011finding, shojaee2013detecting,ott2013negative} and cross-domain~\cite{li2014towards} deceptive opinion spam detection. We use features derived from the LIWC 2015 lexicon, 
which consists of 125 psychologically interesting semantic classes.

\vspace{.05in}
\noindent{\bf POS Features}:
It has been observed that the distribution of 
words' POS tags in a document is indicative of its genre or certain writing style. 
Part-Of-Speech features \cite{biber1999grammar, rayson2001grammatical} have been shown useful for deception detection.

\vspace{.05in}
\noindent{\bf Syntactic Production Rule Features}: \citet{feng2012syntactic} introduced 
syntactic production rule features for deception detection that are drawn from Context Free Grammar (CFG) syntactic parse trees of sentences. We used the Stanford coreNLP tool\footnote{\tt http://stanfordnlp.github.io/CoreNLP/} to obtain both POS tags and CFG production rules for product reviews. In addition, we realized that the bottom 
level syntactic production rules have the form of POS tag $\rightarrow$ WORD and include lexical words, which are dependent on specific domains and their vocabularies overlap with BOWs features.
Therefore, we experiment with the extracted syntactic production rule features with or without the bottom level rules in order to understand their generalities in deception detection, 
and distinguish them as All Production (AP) rules and Unlexicalized Production (UP) rules.  

\subsection{Features Modeling Advertising Language}\label{ad_section}

\noindent{\bf Commonly Used Advertising Phrases}:
The purpose of deceptive review writing is to promote a certain product or to directly persuade customers to buy the product.\footnote{Interestingly, we observed that almost all tasks we crawled from crowdsourcing websites are promoting specific products which might reflect real-life manipulation behavior that promoting is the majority.} Therefore, we hypothesize that deceptive reviews are likely to adopt marketing phrases more frequently than truthful reviews. 
Specifically, we crawled lists of advertising phrases from online blogs\footnote{\tt https://blog.bufferapp.com/words-and-\\phrases-that-convert-ultimate-list} and websites\footnote{\tt http://systemagicmotives.com/Effecti\\ve\%20Ad\%20Words.htm} that provide suggestions for writing persuasive product advertisements. 
The advertising phrases include 
efficient 
phrases to persuade people to make purchase, including cause-and-effect phrases (e.g., due to, thus, accordingly), premium adjectives (e.g., exclusive, guaranteed, unique), premium verbs (e.g., try it, discover, love) and phrases that inspire safety (e.g., authentic, certificated, privacy), etc. Finally, all advertising phrases are used as a set of binary features (presence vs. absence).~\footnote{299 advertising phrases are used in total}

\vspace{0.05in}
\noindent{\bf Ngrams in Product Description Titles}:
We observed that many faked reviews repeatedly mention entire or partial product name. For some specific models or products, they even mention five or more successive words same as in their descriptions.
Reasonably, deceptive review composers have not purchased and used the products, so they more frequently refer back to the product descriptions in order to imagine relevant reviews. In contrast, truthful customers write reviews describing their real experience using the products and rely less on product descriptions.
In specific, we count the number of common features shared by a review and corresponding product description in terms of unigram and bigram and put two frequency scores as new features.


\subsection{Syntactic Complexity Features Indicating Deceptive Writing Styles}\label{syn_comp_section}

Syntactic complexity scores \cite{lu2010automatic} have been shown useful in measuring text readability and distinguishing authorships. These scores have not been used in deception detection. But intuitively, deceptive reviewers do not want to invest too much time to get paid so they unconsciously tend to use simple sentence syntactic structure to write.\footnote{In our dataset, average sentence length is 17.8 for deceptive reviews versus 20.9 for authentic reviews. Average number of clauses per sentence 1.97 for deceptive reviews versus 2.28 for authentic reviews.}
Following \cite{lu2010automatic}, we use a range of measurement scores to represent sentence syntactic complexity, including sentence length, clause length, average number of clauses or specific syntactic constructions per sentence, etc.
Specifically, we use 
the Tregex~\cite{levy2006tregex} system
to query syntactic parse trees\footnote{\tt http://nlp.stanford.edu/software\\/tregex.shtml} using predefined Tregex patterns.

\section{Experimental Results}

\subsection{Data and Settings}

As shown in Table \ref{merge_domain}, we merge similar product categories and create four broad product domains. All the remaining product domains were put under the catch-all ``Other'' category, which is a mixture of a variety of product domains.
Table \ref{domain_size} shows the number of deceptive reviews in each category.

\begin{table}[h]
\small
\centering
\begin{tabular}{|l|l|}
\hline
Large domains                  & Contain categories       \\ \hline
\multirow{3}{*}{Books}         & Hardcover                \\
                               & Paperback                \\
                               & Kindle Edition           \\ \hline
\multirow{2}{*}{Health/Beauty} & Health and Beauty        \\
                               & Health and Personal Care \\ \hline
\multirow{3}{*}{Electronics}   & Electronics              \\
                               & Personal Computers       \\
                               & Cell Phones              \\ \hline
\multirow{2}{*}{Movies}        & Movies and TV            \\
                               & DVD                     \\
\hline
\end{tabular}
\caption{Four Broad Product Domains}
\label{merge_domain}
\end{table}

\begin{table}[h]
\small
\centering
\begin{tabular}{|ccccc|}
\hline
Books & Health & Electronics & Movies & Other \\ \hline
6244 & 2118 & 228 & 292 & 1232 \\ \hline
\end{tabular}
\caption{Number of Reviews for Each Domain}
\label{domain_size}
\end{table}
In the experiments, we use Maximum Entropy~\cite{berger1996maximum} classifiers\footnote{We also tried Support Vector Machines and achieved similar results.}. 
Specifically, we use the implementation of Maxent models in the LIBLINEAR library \cite{Fan2008liblinear} with default parameter settings. 
In reality, there are generally many more authentic reviews than deceptive ones. In order to reflect the actual skewed label distribution, we randomly selected truthful reviews three times of deceptive reviews across our experiments. 
In contrast, the previous studies on deception detection often artificially enforce deceptive and truthful reviews to be balanced. 

\subsection{In-domain Evaluation}
First, we restrict the experiments within one domain where both the training and test data are from one domain. 
Specifically, we conduct in-domain experiments for each of the four broad product domains using 5-fold cross validation. 
We experiment with each of the basic features as described in subsection \ref{basicF}, with one type of features each time.

\begin{table}[h]
\centering
\begin{tabular}{|l|l|}
\hline
Features & Macro Average \\ \hline
unigram   & 80.2/88.0/83.9 \\
POS~\cite{biber1999grammar}  & 32.9/56.7/41.6 \\
LIWC~\cite{pennebaker2015development}  & 46.0/63.4/53.3 \\
AP~\cite{feng2012syntactic} & 76.0/86.0/80.7\\
UP~\cite{feng2012syntactic} & 56.6/62.6/59.5 \\
\hline
\end{tabular}
\caption{In-domain Experimental Results Using Different Features, Recall/Precision/F1-score}
\label{within_eval}
\end{table}

Table \ref{within_eval} shows the macro average scores across four domains when using each type of features. 
Consistent with the previous studies \cite{ott2012estimating,ott2013negative,li2014towards}, the best in-domain performance is achieved using unigrams. In addition, the second best performed type of features is AP features that include the bottom-level syntactic production rules with lexical words. The three other types of features, POS, LIWC and UP features, perform significantly worse on in-domain deception detection. 


\begin{table*}[h]
\centering
\begin{tabular}{|l|lllll|}
\hline
Features & Books & Health & Electronics & Movies & Macro Average \\ \hline
Unigrams                      & 53/73/61 & 19/73/31 & 24/67/35 & 22/79/34 & 29.3/72.9/41.8 \\
POS~\cite{biber1999grammar} & 36/48/41  & 16/52/24 & 22/48/30 & 28/47/35 & 25.5/48.7/33.4 \\
LIWC~\cite{pennebaker2015development} & 35/53/42 & 13/51/21 & 34/46/39 & 23/50/31 & 26.0/50.1/34.2 \\
AP~\cite{feng2012syntactic} & 56/71/{\bf 63}  & 20/74/31 & 27/62/38 & 24/72/36 & 31.7/69.8/43.6 \\
UP~\cite{feng2012syntactic} & 48/55/51 & 32/55/40  & 38/48/43  & 43/48/45 & 40.2/51.5/45.1 \\ \hline
UP + POS                     & 48/55/52 & 32/56/41 & 39/50/44    & 41/48/44 & 40.0/52.3/45.3 \\
\hline
\multicolumn{6}{c}{Results from Neural Net Models} \\ \hline
LSTM~\cite{zaremba2014recurrent} & 45/67/54 & 31/65/42 & 39/57/47 & 34/66/45 & 37.5/63.7/47.2 \\
CNN~\cite{kim2014convolutional} & 45/57/50 & 31/58/40 & 30/52/38 & 41/60/48 & 36.7/57.1/44.7 \\ \hline
\multicolumn{6}{c}{With New Features} \\ \hline
UP + POS + ad & 50/60/55 & 33/63/{\bf 44} & 42/54/{\bf 47} & 43/55/48 & 42.0/58.0/48.8 \\
UP + POS + ad + comp & 51/61/56 & 32/64/43 & 42/54/{\bf 47} & 44/55/{\bf 49} & 42.3/58.5/{\bf 49.1}\\
\hline

\end{tabular}
\caption{Cross-domain Experimental Results Using Different Features, Recall/Precision/F1-score}
\label{cross_eval}
\end{table*}

\begin{table*}[h]
\centering
\begin{tabular}{|l|lllll|}
\hline
Features \hspace{35mm} & Books & Health & Electronics & Movies & Macro Average \\ \hline
UP + POS & 41/62/49 & 40/58/48 & 49/53/51 &47/55/50 & 44.3/56.9/49.8 \\ \hline
\multicolumn{6}{c}{With New Features} \\ \hline
UP + POS + ad & 45/71/55 & 45/66/{\bf 54} & 63/58/60 & 59/60/59 &	52.9/63.6/57.8 \\
UP + POS + ad + comp & 45/73/{\bf 56} & 44/67/53	& 62/60/{\bf 61} & 59/62/{\bf 60} & 52.4/65.4/{\bf 58.2} \\ 
\hline

\end{tabular}
\caption{Cross-domain Experimental Results after Adding Training Data from Other Domains}
\label{cross_eval_with_other}
\end{table*}

\subsection{Cross-domain Evaluation}
To test whether deception detection classifiers implementing general rules perform well across distinct domains, we conduct extensive cross-domain experiments. 
Specifically, with a set of features we train a classifier using reviews from each of four domains -- {\it Books, Health, Electronics, Movies} -- and test the classifier on the rest three domains. 
In each of the four runs, we train a classifier using one domain and report the macro-average recall/precision/F1-score of the classifier across the rest three test domains with respect to deceptive reviews detection. To measure how the set of features performs overall for deception detection, we further calculate the meta macro-average scores over the four sets of macro-average scores resulted from each run.

Table \ref{cross_eval} shows the macro-average scores from each run trained with one domain as well as the meta macro-average scores across the four separate runs.
From the first section of Table \ref{cross_eval}, we can see that in cross-domain experiments 
with each of the five basic types of features, the classifier using UP features outperforms the classifier using Unigrams or AP features overall. Especially in the runs trained with three smaller domains, {\it Health, Electronics, Movies}, UP features are promising in deriving generalized deception detection classifiers. 
Next, we add each of the first three types of basic features on top of UP features. It turns out that POS features slightly improve the overall performance, while both Unigrams and LIWC features hurt the overall performance. 

Table \ref{cross_eval} (With New Features section) shows the cross-domain experimental results when we increasingly add two new types of linguistic features. We can see that the features modeling advertising language can clearly improve the performance of deception detection across the four classifiers trained on each domain, showing that advertising language is commonly seen across four domains. 
Furthermore, adding the syntactic complexity features can slightly improve both the macro-average recall and precision in deception detection.

Long Short-Term Memory (LSTMs)~\cite{hochreiter1997long} and Convolutional Neural Networks (CNNs)~\cite{lecun1998gradient} have been shown effective on deriving compositional meanings of texts and have achieved great success across many NLP tasks
~\cite{sutskever2014sequence, zaremba2014recurrent, kim2014convolutional}. 
For comparison purposes, 
we also conduct cross-domain experiments using both LSTMs and CNNs trained on top of word2vec 300 dimensions word embeddings \cite{mikolov2013distributed}.
\footnote{For LSTMs, we use one hidden-layer of 128 hidden units. For CNNs, we use filter window size 3, 4 and 5, and a hidden-layer of 100 hidden units. For both LSTMs and CNNs, we run Adam optimizer with a learning rate of 0.001, dropout rate of 0.5.}
From the second section of Table \ref{cross_eval}, we can see that the performance of both neural net models are worse than our feature-based classifiers using well selected generalized features. 
One explanation for the lower performance of neural nets on deception detection is that this task is not about understanding semantic meanings of reviews, rather, deception detection is about understanding and recognizing subtle syntactic or stylistic clues and footprints of deceptive writing. Our strong claim is that general rules for deception detection cannot be obtained continuing single-domain studies (in which simple unigram is the best) and we have shown that cross-domain experiments are promising in revealing general rules. 

\subsubsection{Adding Training Data from Distinct Domains}

We have seen that classifiers trained with generalized features perform well across the rest three domains. 
So far, all the classifiers we have used in cross-domain evaluation  
were trained with data from a single domain.  
However, we hypothesize that generalized features should enable deception detection classifiers to further benefit from additional training data, even when the data is from dramatically different domains.

Therefore, we augment each single domain set of reviews with additional reviews from ``Other'' domains (assorted domains) which contain 1,232 deceptive reviews (Table \ref{domain_size}), and rerun the cross-domain experiments. 
\begin{figure}[h]
 \centering
 \includegraphics[width = 2.8in]{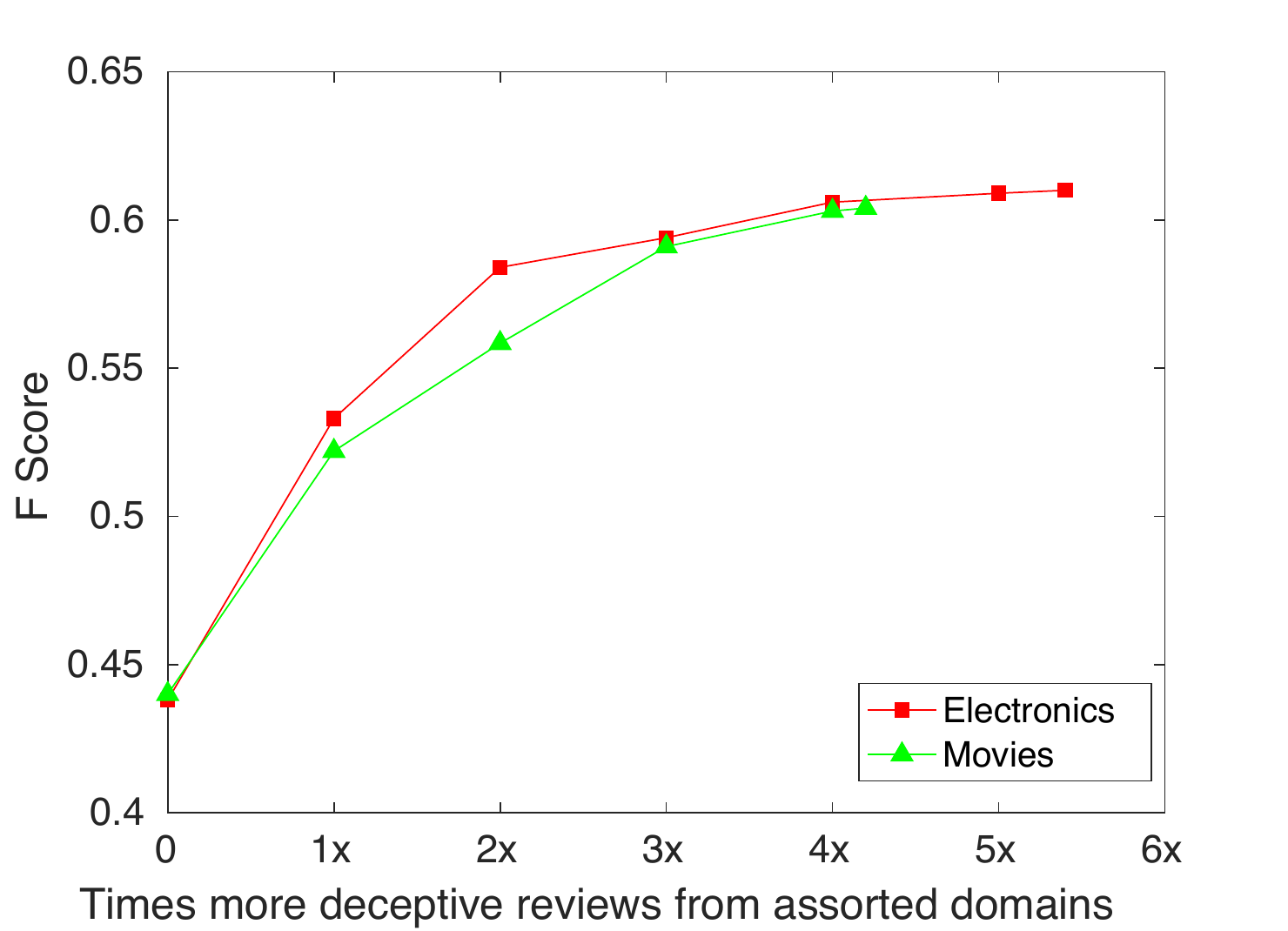}
 \caption{Learning Curves}
\label{learning_curve}
\end{figure}
From Table \ref{cross_eval_with_other}, we can see that using the best set of basic features, UP+POS, the detection performance of classifiers were significantly improved after including additional reviews from assorted domains in training. The improvements are 6-7\% across domains except {\it Books}, where adding 1,232 more deceptive reviews to 6,244 deceptive instances on Books may not notably change its overall review distribution. 
Especially, with the feature sets enriched with our advertising speak features and syntactic complexity features, the performance of the classifier 
even further improves using additional mixed-domain training data, by 10-14\% across the latter three categories (Health, Electronics and Movies). 

Note that for the smallest two domains, {\it Electronics} and {\it Movies} have deceptive reviews of 228 and 292, so the newly added deceptive reviews (1,232) are several times of their original deceptive reviews. In order to understand how the performance of the classifier was influenced when increasingly adding times more training reviews from assorted domains, we drew a learning curve (shown in Figure \ref{learning_curve}) for each classifier that was initially trained with deceptive reviews from one of {\it Electronics} and {\it Movies}. 
We can see that the performance is consistently growing with times more deceptive reviews added in training. We expect to see further improvements if more data were provided. 

These improvements confirm that with generalized features, deception detection performance can be remarkably improved using more data, even with data from dramatically different domains. 
It further emphasizes the value of the new deceptive data collection method that relies on social network analysis to generate amounts of ground truth deceptive reviews across diverse domains.


\section{Effects of Reviewers and Personalized Deceptive Writing Styles}

Intuitively, reviewers of distinct personalities write differently, which 
implies that reviewers should be considered in deriving general rules for online deception detection.
Our dataset includes deceptive reviews that were written by a diverse set of reviewers and many reviewers contributed dozens of deceptive reviews, 
which enables us to study the effects of reviewers in deception detection. 
In the following, we present our initial findings.

\begin{table}[h]
\small
\centering
\begin{tabular}{|cccc|}
\hline
Reviewer 1 & Reviewer 2 & Reviewer 3 & Reviewer 4 \\ \hline
113 & 112 & 78 & 80 \\ \hline
\end{tabular}
\caption{Number of Reviews for Each Person}
\label{num_person}
\end{table}

\begin{figure}[h]
 \centering
 \includegraphics[width = 2.2in]{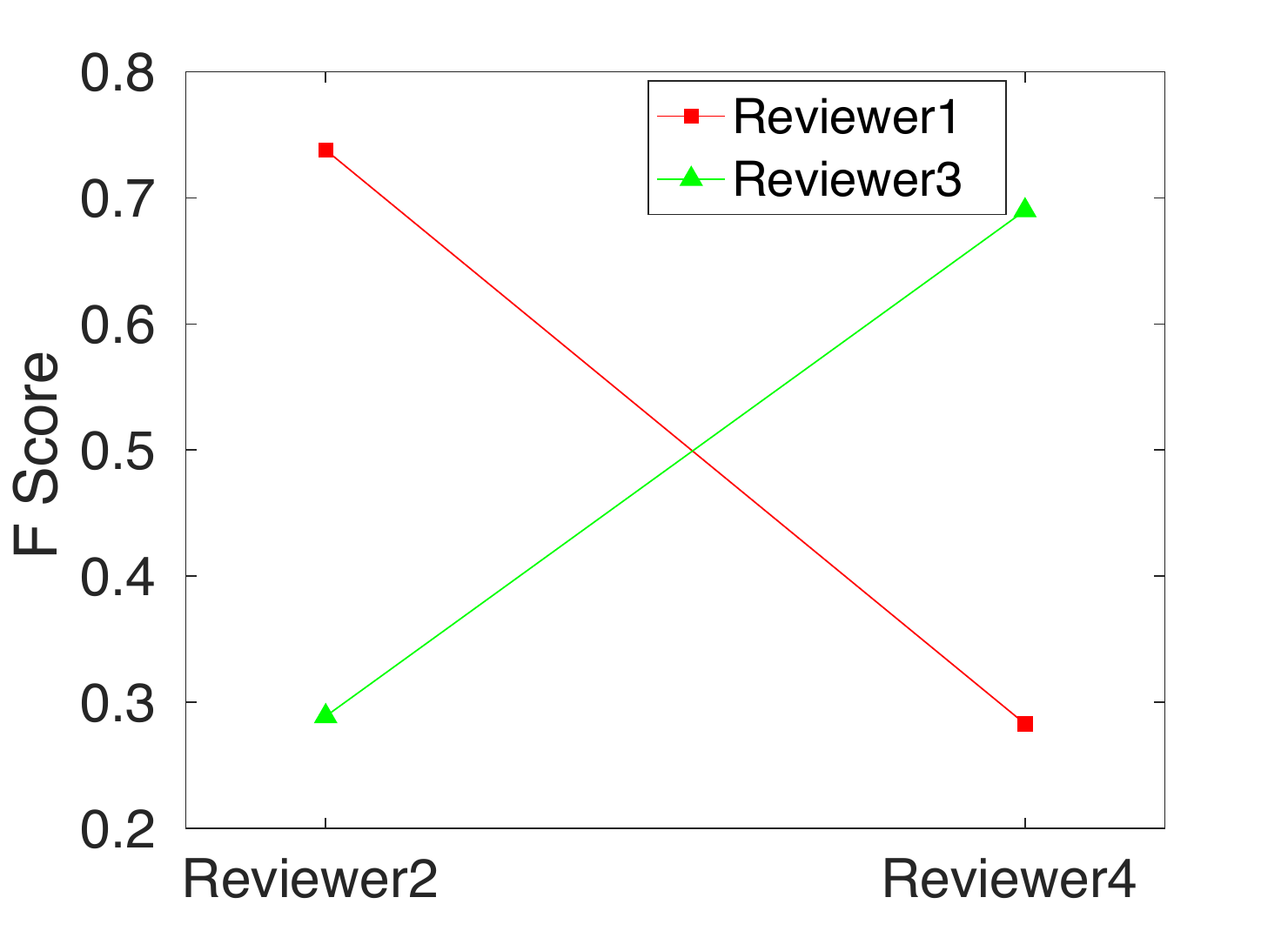}
 \caption{Reviewer level evaluation}
\label{reviewer_eval}
\end{figure}

We identified four deceptive reviewers that fall into two groups based on their writing styles (Table~\ref{num_person} shows the number of deceptive reviews for each reviewer). 
Specifically, we observed that reviews by the first group of reviewers use shorter  sentences and advertising words more frequently in making comparisons. In contrast, reviews by the second group of reviewers tend to use longer and generally more complex sentences, more numbers and more words related to personal feelings.
Figure \ref{reviewer_eval} shows the result when we train the classifier using reviews from one reviewer taken from each category, say reviewer R1 and R3, and apply the classifier to deceptive reviews written by the other two reviewers, R2 and R4, also one from each category. The detection performance can achieve F-scores as high as 70-80\% when it is trained and tested on 
the same type of reviewers. In contrast, detection F-scores can be as low as 30\% when it is trained on one type and tested on another type. Deceptive examples from each reviewer is also listed in Table~\ref{review_example}.

\begin{table}[htbp]
\small
\centering
\begin{tabular}[center]{|l|} \hline
{\it Reviewer 1:} LOVE TO BAKE. I love these type of recipes \\ especially for cupcakes and cakes! I have tried almost them \\all so far. It's the best book for finding that perfect flavor for \\that perfect event your planning. Highly recommended! \\
\hline
{\it Reviewer 2:}
A cute kids book! I was honestly impressed\\ with this childrens book. I read it to my son and it kept his\\ interest, which is no easy task! He liked the front cover\\ picture. This was a hit with my son. Great book!
\\
\hline
\hline
{\it Reviewer 3:} Very Valuable Information. The reason\\ why I bought this book is because I really needed a boost\\ in self confidence, my main luck of confidence was in\\ social situation where I tended to shy away and keep quite,\\ I must say that after reading this book my confidence level\\ went up and I feel much more comfortable when I am out\\ with friends, I now talk more, engage more in the\\ conversations and feel much better when out with friends. \\
\hline
{\it Reviewer 4:} Diet is awesome. Well..you know life is hard\\ when you're fat, but if you get this you may succeed in\\ your diet and you will lose some pounds. My grandmother\\ was fat i bought this book for her and she is pretty well\\ now if you want to lose your weight you may seriously\\ want to try this book. I recommend it to anyone congrats\\ to the maker and good luck selling more copies.
\\
\hline
\hline
\end{tabular}
\caption{One Example Deceptive Review per Reviewer}
\label{review_example}
\end{table}

\vspace{-0.1in}

\section{Conclusion}
We applied a new method for real world deceptive review collection leveraging social network analysis. The newly collected dataset is rich in products and reviews and has two orders of magnitude more deceptive reviews than previously used artificial datasets.
We demonstrate that such a large dataset facilitates the development of identifying generalized features for deception detection. We further show that with generalized features, additional deceptive review data from assorted domains can be used to improve both recall and precision of online deception detection. 




\bibliography{acl2017}
\bibliographystyle{acl_natbib}

\end{document}